\documentclass[conference]{IEEEtran}
\IEEEoverridecommandlockouts
 
\usepackage{cite}
\usepackage{amsmath,amssymb,amsfonts}
\usepackage{mathtools}
\usepackage{algorithm}
\usepackage{algpseudocode}
\usepackage{graphicx}
\usepackage{booktabs}
\usepackage{multirow}
\usepackage{subcaption}
\usepackage{xcolor}
\usepackage{siunitx}
\usepackage{balance}
\usepackage[hidelinks]{hyperref}
\usepackage{cleveref}
\usepackage{comment}
\usepackage{bm}

\DeclareMathOperator*{\argmax}{arg\,max}
\usepackage{colortbl}
\newcommand{\bst}[1]{\cellcolor{green!22}\textbf{#1}}   
\newcommand{\snd}[1]{\cellcolor{blue!13}#1}             
\newcommand{\gm}[1]{\textcolor{black!55}{#1}}           
\newcommand{\ppm}[1]{\,\raisebox{0.15ex}{\tiny$\pm$#1}}

\newcommand{\err}[1]{%
  \ifnum#1>27 \cellcolor{red!60}\else
  \ifnum#1>18 \cellcolor{red!42}\else
  \ifnum#1>9  \cellcolor{red!24}\else
  \ifnum#1>0  \cellcolor{red!10}\fi\fi\fi\fi #1}

\graphicspath{{figures/}}
 
\begin{document}

\title{\vspace*{12pt}Concurrent Semantic Search and Mission Execution for LTL
Missions in Unknown Environments}
 
\author{Fernando Salanova*, David Morilla, Cristian Mahulea, Eduardo Montijano
\vspace*{-12pt}{\thanks{Disclosure of AI usage: Anthropic-Claude models were used for
assisting in code development of the method.}}}
 
\maketitle
\thispagestyle{empty}
\pagestyle{empty}

\begin{abstract}

Planning complex missions in unknown environments requires robots to reason simultaneously about what they should do and what they still need to discover. Existing approaches for solving LTLf missions typically assume a known environment, or separate the exploration of the environment from the execution of the mission, while semantic exploration methods look for one target at a time and ignore the mission being executed.

To fill this gap, our main contribution is an adaptive high-level planning method that interleaves a task-driven semantic search with the execution of the mission, advancing both in a non-myopic manner.

Our method leverages two representations built online, a metric-semantic scene graph, built with a Vision Language Model (VLM), that provides the evidence needed to locate the objects the mission refers to, and the deterministic finite automaton (DFA) encoding the mission, that indicates which of them matter at each mission state. At every planning stage, our planner selects the waypoints that are most valuable for both the semantic search and the advancement of the
mission, valuing them over the remaining mission stages in order to avoid blocking states. The selected waypoints are then ordered in a single high-level plan, which is recomputed as new information arrives.

In photorealistic indoor environments over five mission types, our method completes more missions than the compared approaches while having to cover less of the environment, and it does so with shorter paths and complying with the restrictions imposed by the mission.

---Anonymized code implementation: \url{https://anonymous.4open.science/r/Co-Explore-LTL-E2F4}

\end{abstract}

\section{Introduction}
\label{sec:intro}

\begin{figure}[t]
  \centering
  \includegraphics[width=\columnwidth]{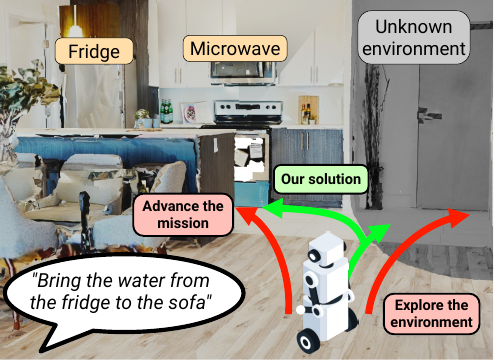}
  \caption{We address the problem of solving LTLf missions in unknown environments, which requires exploring the scene to find the objects the mission refers to. Our adaptive planning method interleaves both objectives, finding paths that gather the information the mission requires while advancing the stages that can already be satisfied.}
  \label{fig:teaser}
\end{figure}

Robots are becoming more prevalent for performing complex autonomous tasks in changing or unknown environments. Linear temporal logic (LTL) enables formulating high-level missions involving actions related to semantic object categories, with temporal and logical constraints between them~\cite{vardi2005automata}. These missions can be compiled into different representations that allow planners to generate compliant paths~\cite{luo2022temporal}. However, in unknown environments, robots must first acquire information about where the objects related to the mission are, Fig.~\ref{fig:teaser} shows an overview of our proposed solution for this problem.

Existing approaches for solving LTL missions in unknown environments usually separate the exploration from the mission execution, exploring the environment in a first stage until there is enough information to complete the mission, and then computing a plan to complete it~\cite{lamanna2021online,quartey2025verifiably}. However, this separation is inefficient, since the exploration is agnostic to the mission and requires redundant navigation and revisiting already explored
regions. On the other hand, combining exploration with mission advancement in a myopic way, considering only the next mission stages, might result in blocking states, i.e., mission states from which the mission can no longer be satisfied, for example due to the absence of an object required for the
mission~\cite{kantaros2022perception,taheri2026temporal}.

The recent emergence of Vision-Language Models (VLMs) has enabled extracting rich semantic information from the images captured by the robot and relating it to a mission expressed in natural language~\cite{liu2024grounding,gu2024conceptgraphs}. Furthermore, their reasoning ability enables identifying the regions where the objects of interest might be located from the surrounding semantic
context~\cite{zhou2023esc,yu2023l3mvn,yokoyama2024vlfm}. This offers the possibility to build a metric-semantic representation of the environment and to extract the relationships between the LTL mission and the object categories it contains. In this spirit, the robot can perform a mission-driven exploration, directing its observations towards the regions where the mission is likely to be advanced instead of exploring the environment exhaustively.

In this paper, we target the problem of solving LTL missions in unknown environments efficiently. We propose a novel adaptive high-level planning method to interleave a task-driven semantic search with the execution of the mission. First, instead of exploring the environment exhaustively, our method uses a VLM to build an environment representation and guide the semantic search. Then, considering the current information about the scene and the state of the mission, our planner finds a sequence of waypoints that jointly gathers missing information for the mission while advancing the stages that can be satisfied. The plan is recomputed as new information arrives, and the remaining mission stages are taken into account to avoid blocking states. We evaluate our method in photorealistic indoor environments over several LTLf missions compared to existing planning methods.The results show that our method completes more missions than the compared approaches, while having to cover less of the environment thanks to the targeted exploration. This results in a higher efficiency in mission time and traversed distance, and in complying with the restrictions imposed by the mission. We provide the implementation of our method for anonymous review at \url{https://anonymous.4open.science/r/Co-Explore-LTL-E2F4} and we will release it publicly upon acceptance.

\section{Related Work}
\label{sec:related}

\subsection{Temporal-logic planning in unknown environments}

Temporal-logic planning is classically posed over a known workspace with a known labeling, the formula is compiled into an automaton, composed with an abstraction of the environment, and a satisfying path is synthesized before the robot moves~\cite{kloetzer2009automatic,luo2022temporal}. Later work removes parts of that knowledge, reconfiguring the plan as the map is revised~\cite{guo2015multi, lahijanian2016iterative,kantaros2020reactive}, planning over frontiers whose approach would not violate the task~\cite{ayala2013temporal}, guiding the planning by task progress~\cite{grover2021semantic}, and planning over maps with uncertain labels~\cite{kantaros2022perception,kalluraya2022multi}. The automaton has also been used to steer exploration away from the states that discard accepting behaviors~\cite{taheri2026temporal}.

In these methods exploration is not a part of the works, so they abstract information gathering or use geometric exploration without any semantic guidance.

\subsection{Semantic exploration}

Exploration is classically driven by geometry, selecting frontiers between explored and unknown space~\cite{yamauchi1997frontier,zhou2021fuel}. Semantic evidence reduces the redundant navigation this produces, since objects already seen indicate where others are likely to be~\cite{aydemir2013active,park2023zero,hanheide2017robot}.
Vision-language models supply that evidence without a fixed set of classes~\cite{liu2024grounding,gu2024conceptgraphs,werby2024hierarchical,liu2024fm}, and object-goal navigation uses it to explore until a requested category is detected~\cite{chaplot2020object,yokoyama2024vlfm,zhou2023esc,yu2023l3mvn}.

The target of these guided search are single objects. For a temporally structured mission the relevance of an observation is not fixed, needing further complexity.

\subsection{Exploring and executing a mission}

Solving the complete problem requires both. The common approach separates them sequentially, exploring the environment first and planning over the result afterwards~\cite{lamanna2021online,quartey2025verifiably}, which causes revisiting. Planning over partial information instead requires the execution to be non-myopic, so that a step taken now does not prevent the mission from being completed later~\cite{kantaros2022perception,taheri2026temporal}. Exploration and task actions have also been scored jointly~\cite{xu2022framework,yang2026epog,tzes2022reactive}, although over a planning domain that cannot express order, avoidance or disjunction.

We address the complete problem with both at once, exploration is mission driven, and execution is non-myopic performing both concurrently.

\begin{figure*}[t]
  \centering
  \includegraphics[width=\textwidth]{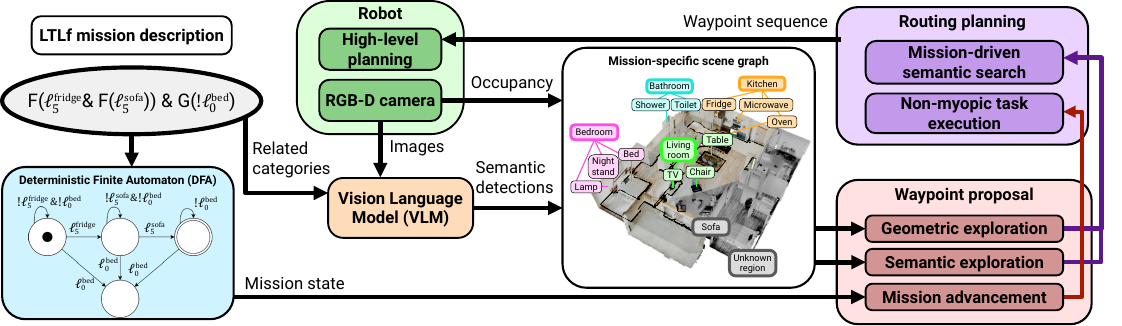}
  \caption{Overview of the proposed method. The LTLf mission is compiled into a DFA that
tracks its state and provides the categories related to the mission objects. The
VLM build the scene graph, from which three kinds of waypoint are proposed, geometric exploration, semantic exploration and mission advancement. The planner returns the next
waypoint of the path based on value and distance, repeating as the scene graph grows and the mission advances.}
  \label{fig:diagram}
\end{figure*}

\section{Problem Formulation}
\label{sec:problem}

Consider a robot operating in a bounded and unknown indoor environment whose pose at discrete time steps $k$ is denoted as $p_k \in \mathbb{R}^2$.
The robot is tasked with a mission specified as an LTLf formula $\varphi$ over a finite set of atomic propositions, $AP$.
Each proposition, $\ell^\lambda_\tau \in AP$, represents an action associated with an object category $\lambda$ (e.g., fridge, sofa) and a required execution time $\tau$. We consider that object categories are defined via natural language, allowing arbitrary categories in the mission.
Propositions are evaluated \emph{true} if the robot remains within the proximity of the associated object $\lambda$ for the specified time $\tau$, and \emph{false} otherwise.
This compact representation enables describing household tasks by composing object-centered actions with temporal and logical constraints. For example, the task of \emph{``taking the sheets from the washing machine and changing the bed without bothering people on the sofa''} can be specified as
\[
\varphi := \mathbf{F}\left(\ell^{\text{washing machine}}_{10} \wedge \mathbf{F}\, \ell^{\text{bed}}_{90}\right) \wedge \mathbf{G}(\lnot \ell^{\text{sofa}}_0),
\]
which requires the robot to first interact with the washing machine for 10 time units and subsequently interact with the bed for 90 time units without interacting with the sofa at any time.
In this paper, we assume that underlying actions associated with each object are solved by low-level control strategies, focusing on the high-level planning problem.

Since we consider an initially unknown environment, the problem cannot be solved before exploring the scene to locate objects present in $\varphi$. Furthermore, advancing the mission on partial knowledge myopically can lead to blocking situations. Consequently, the robot must use its sensors to acquire relevant information about the environment that enables satisfaction of the mission.

We therefore address the problem of planning to satisfy an LTLf-defined mission over an unknown environment, requiring a high-level plan to (1) explore the unknown environment to find objects related to the mission, and (2) execute the mission efficiently in a non-myopic manner to avoid blocking situations. Rather than separating exploration from task execution into separate stages, our objective is to develop a unified solution that addresses both problems simultaneously in an informed manner.

\section{Methodology} \label{sec:method}

We propose a novel adaptive high-level planning method to tackle the problem of
task-driven information acquisition and mission execution depicted in Fig.~\ref{fig:diagram}. We incrementally
build a metric-semantic scene representation in the form of a scene graph,
$\mathcal{S}_k,$ using robot observations from an RGB-D camera at keyframes and
a Vision Language Model (VLM) to extract semantic concepts
(Sec.~\ref{sec:map}). We also monitor the state of the LTLf mission towards its
satisfaction, denoted by $q_k$, at every time step $k$ via a deterministic finite automaton (DFA),
$\mathcal{A}_\varphi,$
(Sec.~\ref{sec:mission}). Our solution leverages these two representations to
jointly perform a task-driven semantic search while efficiently advancing the
mission in a non-myopic manner to avoid blocking states. At each replanning stage $t_k$, we aim at finding the
ordered sequence of robot 2D high-level waypoints,
$\pi_k=\left(w_1,\ldots,w_N\right)$, that solves
\begin{equation}
  \pi_k^* = \argmax_{\pi_k}
  \bigl[\, V(\mathcal{S}_k,\mathcal{A}_\varphi,q_k,\pi_k) - C(\pi_k) \,\bigr],
  \label{eq:objective}
\end{equation}
where $V$ is a value function considering both information acquisition and mission
advancement, and $C$ is the cost of navigating the waypoints, which is determined
by their order in the plan. As $\mathcal{S}_k$ and $q_k$ vary over
planning stages, the high-level plan is updated to adapt over the mission
considering new available information.The planning horizon $N$ is given by everything that can be planned
with the information available at $t_k$.

We solve \eqref{eq:objective} in two stages. First, we search for the set of
relevant waypoints $\mathcal{W}_k$ to visit that maximizes the value of the plan, $V,$ by
intelligently exploring the environment while advancing the mission
(Sec.~\ref{sec:candidates}). Then we weigh those values against the travel each waypoint demands and select the best
(Sec.~\ref{sec:cost}). This process is repeated until satisfaction of $\varphi$ or, if no feasible continuation exists and the mission can no longer be fulfilled, the robot stops.

\subsection{Scene representation} \label{sec:map}

Since the environment is not known, a scene representation is required to enable spatial and semantic awareness for the mission execution in terms of navigation, fulfillment of propositions and targeted semantic exploration.
We therefore incrementally construct a scene graph, $\mathcal{S}_k,$ at each discrete time step k, integrating metric and semantic information.

Geometric information is obtained from the robot sensors such as a LiDAR or depth camera. Range measurements are integrated in a 2D probabilistic occupancy gridmap $\mathcal{M}_k$ which constitutes the base layer of the scene graph. Each cell is classified as free, occupied, or unknown. The geometric layer is used to determine navigable space, compute paths between locations, and identify the boundary between already explored and unknown regions.

Semantic information is required to identify object categories $\lambda$ corresponding to APs in the $\varphi$. Let $\mathcal{L}_\varphi=\{\lambda \;|\; \exists \tau \in \mathbb{R}_{\geq 0} \text{ such that  } \exists \ell^\lambda_\tau \in AP\}$ the set of objects specified in the LTLf mission. Additionally, we want to identify a set of spatially correlated categories, denoted as $\mathcal{L}_{rel}$ to drive the semantic search towards potentially informative regions. For example, if the robot needs to visit the \emph{fridge}, then we could expect \emph{microwave} and \emph{kitchen} present in $\mathcal{L}_{rel}.$

In order to acquire this semantic understanding, we leverage the reasoning capabilities of a VLM. Prior to the mission, we query the VLM to obtain the set of related categories $\mathcal{L}_{rel}$ for $\mathcal{L}_\varphi.$ During the mission, we capture RGB frames from the robot camera at keyframe locations with sufficient spatial variability between them. We then query the VLM to suggest a room category for the global image, denoted by $h$, to identify an unconstrained list of object categories, which grounds the detections in their context and reduces false positives, their position on the frame, and an estimated confidence for each detection. The object detections are projected to the metric map $\mathcal{M}_k$ using the range sensor. Every detection is stored as an object prototype, $o$, which constitute the second layer of our scene graph, $\mathcal{O}_k.$ An object node $o \in \mathcal{O}_k$ is then represented as
 \begin{equation}
  o=(\lambda_o,x_o,c_o,n_o,h),
  \label{eq:instance}
\end{equation}
where $\lambda_o$ is the predicted class, $x_o$ its 2D position in $\mathcal{M}_k,$ $c_o$ the VLM confidence in the detection, $n_o$ the number of times detected (initially $1$) and $h$ the room where it was seen.
When a projected object detection is near an already existing prototype, they are fused averaging their spatial locations, keeping the maximum confidence and incrementing a track of observations for that object.

Finally, we cluster by distance the room categories associated to objects to build a third layer for room categories, $\mathcal{H}_k$. The room extent is determined by the cluster size, and its position is given by the cluster center, $x_h$. Therefore the scene graph construction is defined as $\mathcal{S}_k=\left(\mathcal{M}_k, \mathcal{O}_k, \mathcal{H}_k \right).$

We use $\mathcal{S}_k$ to detect regions related to $\varphi$ by identifying the object nodes in $\mathcal{L}_\varphi,$ and semantically interesting regions to explore in search of mission objects by identifying object and room nodes in $\mathcal{L}_{rel}$

\subsection{Deterministic Finite Automaton} \label{sec:mission} 

To track the state of the mission, $\varphi$ is compiled by the translation~\cite{kupferman2001model} into a deterministic finite automaton,
\begin{equation}
  \mathcal{A}_{\varphi}=(Q,q_0,\Sigma,\delta,Q_F),
  \label{eq:dfa}
\end{equation}
whose states, $Q$, represent different stages of the mission, $q_0$ is the initial state, and $Q_F$ the states at which $\varphi$ is satisfied. We consider missions whose automaton is acyclic, so progress cannot be undone.
At every instant $k$, each proposition $\ell^\lambda_\tau\in AP$ is either true or false. 
The automaton describes this with the alphabet $\Sigma=\{0,1\}^{|AP|}$, where the entry $\sigma[\ell^\lambda_\tau]$ of a symbol $\sigma\in\Sigma$ is $1$ when $\ell^\lambda_\tau$ is true and $0$ otherwise. Each symbol labels a transition $\delta:Q\times\Sigma\rightarrow Q$, so the state of the mission changes according to which propositions hold, $q_{k+1}=\delta(q_k,\sigma_k)$.

Since the ground truth value of $\sigma$ is not available to the robot, due to the lack of precise information about the objects, our solution resorts to estimating it at each time from $\mathcal{S}_k.$

\subsection{Waypoint definition} \label{sec:candidates}

From the partial environment representation $\mathcal{S}_k$ and the current state of the mission $q_k$, we must now find the next waypoints $\mathcal{W}_k$ for the robot to concurrently acquire relevant information and advance the mission. To narrow the search space, we sample a finite set of waypoints, $\mathcal{W}_k = \left\{w_0, \ldots, w_N\right\},$ at interesting locations.

\textbf{Geometric exploration}, $\mathcal{W}_g \subseteq \mathcal{W}_k$. We sample waypoints $w_g$ at the frontiers of $\mathcal{M}_k$, free regions of the geometric layer neighboring unknown ones, to observe new regions of the space potentially containing relevant information.

\textbf{Semantic exploration}, $\mathcal{W}_s \subseteq \mathcal{W}_k$. Object and room categories in $\mathcal{L}_{rel}$ indicate where the objects of $\mathcal{L}_\varphi$ pending to be found are likely to be. Therefore, we identify in $\mathcal{S}_k$ the position of relevant objects and rooms. Categories related to the same $\lambda\in\mathcal{L}_\varphi$ lying within a radius $r_c$ are grouped into a cluster $\mathcal{C}_{w_s}$. Then, a semantic exploration waypoint is placed at its confidence-weighted centroid,
\begin{equation}
  w_s=\frac{\sum_{o\in\mathcal{C}} c_o\,x_o}{\sum_{o\in\mathcal{C}} c_o},
  \label{eq:cluster}
\end{equation}
towards the orientation observing most of the cluster items. Rooms generate a $w_s$ at their center positions.

\textbf{Mission advancement}, $\mathcal{W}_m \subseteq \mathcal{W}_k$. Once an object category associated with a proposition $\ell^\lambda_\tau$ that appears on an outgoing transition of the current DFA state $q_k$ is identified in $\mathcal{S}_k$, we place a waypoint $w_m$ next to the object to advance $q_k$ in $\mathcal{A}_\varphi$.


 \begin{figure}[t]
  \centering
  \includegraphics[width=\linewidth]{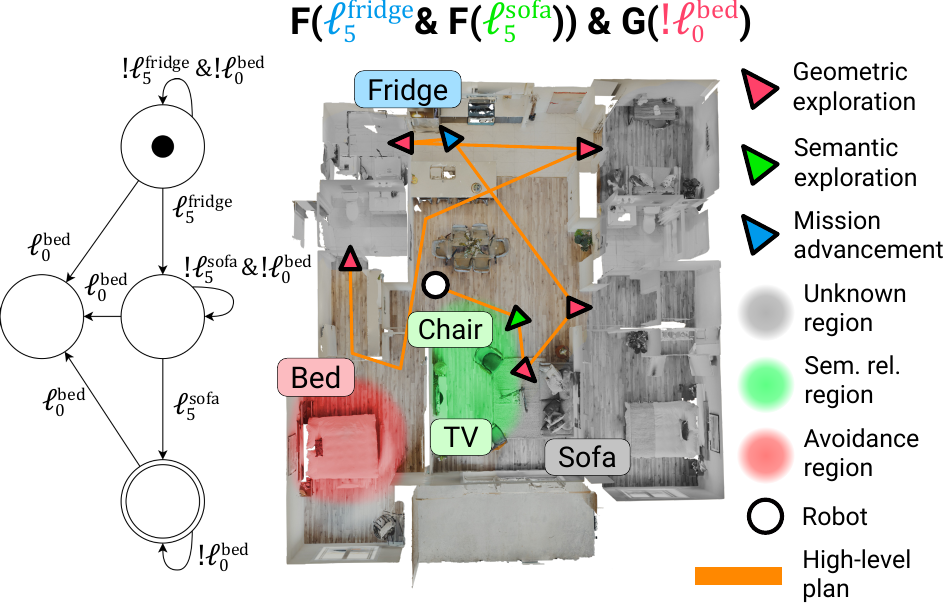}
  \caption{A planning stage for the mission above. The waypoints are sampled from the scene graph $\mathcal{S}_k$ and the mission state $q_k$. The fridge is the proposition to reach, generating a mission advancement waypoint; the chair and the TV are related to the sofa, required later, so their cluster generates a semantic exploration waypoint; additional waypoints lie at the frontiers of the unexplored space. The bed is forbidden, and its region is avoided. The waypoint values are balanced with their relative distance to compute the high-level plan. As the robot advances, the waypoints and the plan are computed again.}
  \label{fig:planning}
  \vspace{-15pt}
\end{figure}

\subsection{Planning for exploration \& mission advancement} \label{sec:cost} 

Given the set of candidate waypoints $\mathcal{W}_k = \mathcal{W}_g \cup \mathcal{W}_s \cup \mathcal{W}_m$, we must now decide the high-level plan to visit them. Our planner considers how much each waypoint contributes to the mission, which we express as its value $v(w)$, and how far it is from the robot, which determines the navigation cost.

\textbf{Exploration value.} Waypoints in $\mathcal{W}_g$ and $\mathcal{W}_s$ contribute by directing the exploration towards where the relevant objects for the mission might be. Since a geometric waypoint, $w_g$ is not semantically informed, we assigned a low base value $v_g$.
A semantic waypoint, $w_s,$ generated by cluster $\mathcal{C}_{w_s}$ is instead valued according to the semantic evidence that its members provide,
\begin{equation}
  v_s(w) = \frac{1}{n_{\mathrm{ref}}}
  \min\Bigl(n_{\mathrm{ref}},\;\textstyle\sum_{o\in\mathcal{C}_{w_s}}c_o\Bigr),
  \label{eq:vsem}
\end{equation}
where $c_o$ is the confidence of the object $o$ and $n_{\mathrm{ref}}$ is the accumulated confidence at which the value saturates at $v_s^{\max}$. In this
way, a cluster supported by a single uncertain object barely provides value,
whereas a cluster containing several related objects is interesting for the
mission. Room semantic waypoints are given a constant value $v_h$. Once the
robot has visited a cluster without finding the related object, the value of its
waypoint is set to zero for the remainder of the mission.

\textbf{Mission advancement value.} For valuing a waypoint $w_m$, we examine the outgoing transitions from the current DFA state $q_k$ associated with that waypoint. For each such transition, we denote its successor state by $q'$ and consider the possible continuations from $q'$ to the accepting states $Q_F$.
We say that another state $q''$ is reachable from $q'$ if there is a path in $\mathcal{A}_\varphi$ using propositions already instantiated in $\mathcal{S}_k$.

Moreover, we denote by $r(q')$ the minimum number of transitions from $q'$ to $Q_F,$ which is set to $\infty$ in case no path exists,
and $b(q')$ the number of possible paths in the DFA to reach $Q_F$ from $q'$. 
The first function is useful to quantify the potential cost to reach $Q_F$ whereas the second one helps us identify potential deadlocks caused by unseen objects in branching missions.
The value is then
\begin{equation}
  v(q')=
  \begin{cases}
    v_{\max}-\beta\,r(q'), &
    b(q')=1 \text{ or } \\
    & \kern -55pt \exists q''\text{ reachable from } q'\text{ such that } b(q'')=1,\\[2pt]
    -\Lambda, & \text{otherwise.}
  \end{cases}
  \label{eq:vmis}
\end{equation}

Waypoint $w_m$ may fire more than one of the transitions, since several propositions can hold. We therefore denote by $Q'(w)$ the successor states of all the transitions associated with $w$, and take the minimum of them to be conservative and account for the worst case scenario,
\begin{equation}
  v_m(w)=\min_{q'\in Q'(w)} v(q').
  \label{eq:vmin}
\end{equation}

Notice that this valuation is non-myopic, as it is read from the whole automaton and not only from the transition ahead, so each waypoint is valued knowing how much of the mission would remain after reaching it. 

\textbf{Navigation cost and plan.} Navigating towards a waypoint requires a cost related to the travel to it. We compute $d(w_i, w_j)$ as the length of the shortest free path in $\mathcal{M}_k$ between the two waypoints, taking the
current robot pose $p_k$ as the origin $w_0$ of the plan. We then bias this cost with the value provided by the waypoint,
\begin{equation}
  c(w_i,w_j)=d(w_i,w_j)-\gamma\,v(w_j),
  \label{eq:cost}
\end{equation}
where $v(w_j)$ is the value assigned above according to the type of waypoint, and $\gamma$ translates values into distances, meaning that a value $v$ is worth a detour of $\gamma v$.

Every waypoint in $\mathcal{W}_k = \mathcal{W}_g \cup \mathcal{W}_s \cup \mathcal{W}_m$ is included in the plan, which we build greedily by appending, at each step, the waypoint balancing cost and value from the last one,
\begin{equation}
  w_{\pi_{j+1}} = \arg\min_{w\in\mathcal{W}_k\setminus\pi_j}
  \left[ d(w_{\pi_j},w)-\gamma\,v(w) \right],
  \quad w_{\pi_0}=p_k,
  \label{eq:select}
\end{equation}
where $\pi_j$ is the sequence of waypoints already appended. The waypoint executed next is the first element of the plan, $\pi_k^*=w_{\pi_1}$. While the robot heads towards it, \eqref{eq:select} is solved again with the current $\mathcal{S}_k$ and $q_k$, which makes the planner adaptive and changes its course whenever another waypoint becomes the better
choice. Selecting the waypoints in this way does not make the plan myopic, as their value is read from the whole automaton and not only from the transition ahead. Exploring and advancing the mission are therefore never alternated, both are waypoints of $\mathcal{W}_k$, and one solution of \eqref{eq:select} serves both. Fig.~\ref{fig:planning} exemplifies with an illustration a cycle of the planner.

\section{Implementation details}\label{sec:imp}

\textbf{Mapping.} The occupancy layer is built with slam\_toolbox at \SI{0.05}{\metre} per cell from a 2D lidar with a \SI{110}{\degree} forward field of view and a range of \SI{4}{\metre}.
 
\textbf{Semantic layers.} RGB-D keyframes are captured every \SI{0.15}{\metre} or \SI{10}{\degree}, at most every \SI{0.3}{\second}, dropping blurred and redundant ones. Each is analyzed by Qwen3.5-9B, which returns the household objects in view as open-vocabulary nouns with a bounding box and a confidence, together with the room type. Detections below a confidence of $0.65$ are discarded, and the rest are projected using the depth measurements and merged with same-category instances within \SI{2.5}{\metre}. 

Rooms group keyframes of the same type within \SI{3}{\metre}, and an object belongs to the room holding the most samples within \SI{2.5}{\metre} of it.

\textbf{Mission monitoring.} A proposition holds within $r_{\mathrm{visit}}=\SI{0.75}{\metre}$ of an instance of its category after $\tau_{\mathrm{visit}}=\SI{5}{\second}$, with $\tau_{\mathrm{visit}}=0$ under negation, so a constraint is violated immediately when entering its range. The mission is declared infeasible when no frontier remains and a bounded search over the automaton, restricted to the categories already mapped, finds no way to complete it.

\textbf{Values.} All waypoint values are designed and normalized to lie in $[0,1]$. We use $v_g=0.1$ for a frontier, $v_h=0.4$ for related rooms, $\min(1,\sum c/n_{ref})$ with $n_{ref}=3$ for a cluster, $\mathcal{C}_{w_s}$ of related instances. The penalty of \eqref{eq:vmis} is $\beta=0.15/r_{\max}$, so a way of finishing that requires
every remaining proposition loses $0.15$, and $\Lambda=50$ places a waypoint behind all the others.
Their influence is bounded by $\gamma$, which converts value into distance, with $\gamma=\SI{4}{\metre}$ no difference in value sends the robot more than four meters out of its way, so values order waypoints of comparable distance rather than override the geometry. The same values are used for every scene and mission
type, without tuning.

\textbf{Planning.} The loop runs at \SI{2}{\hertz}, re-solving the order every cycle.
 
\textbf{Platform.} Simulated turtlebot robot in Gazebo Harmonic under ROS~2 Jazzy on Ubuntu 24.04, with a differential-drive Kobuki base limited to \SI{0.6}{\metre\per\second} and \SI{1.5}{\radian\per\second}, driven with Nav2 and an MPPI controller.


\section{Experiments Setup}
\label{sec:exp}

The experiments evaluate whether reasoning jointly about exploration of the environment and the current state of the mission execution, and following a task-driven semantic search (i) improves the completion of LTLf missions and (ii) yields more efficient plans than separating the stages or considering a single step in the future. 

\subsection{Setup}

Experiments are run in HM3D-Sem evaluation set~\cite{yadav2023habitat}, which provides 36 photorealistic household reconstructions with annotated object categories. An episode is a scene, an initial pose $p_0$ and a mission $\varphi$, started with no prior occupancy or semantic map. Each episode ends when the mission is satisfied, violated or exploration finishes without finding all the objects in $\mathcal{L}_\varphi$. Success and errors are measured against the ground-truth object positions, while the execution plans over the estimates in $\mathcal{S}_k$.

We evaluate several mission families listed in Table~\ref{tab:missions} and ordered by behavior type within LTLf. Each $\ell^{\lambda}_{\tau}$ is the proposition associated with an object $\lambda$, instantiated with different categories in each episode, with $\tau=\SI{5}{\second}$ for propositions to be satisfied and $\tau=0$ for those under negation.

Each mission type is evaluated over four scenes, with three variations of the formula and three initial poses per scene, resulting in 36 episodes per mission type and 180 in total. Simulations run on Ubuntu 24.04 with an AMD Ryzen 9 9950X, 64~GB of RAM and an NVIDIA RTX 5090.

\begin{table}[!t]
\centering
\caption{LTLf mission families used in the experiments.}
\label{tab:missions}
\small
\setlength{\tabcolsep}{4pt}
\begin{tabular}{@{}ll@{}}
\toprule
Mission & LTLf formula \\
\midrule
Object search & $\mathbf{F}\ell^{a}_{5}$ \\[2pt]
Sequential & $\mathbf{F}\bigl(\ell^{a}_{5}\wedge\mathbf{F}(\ell^{b}_{5}\wedge\mathbf{F}\ell^{c}_{5})\bigr)$ \\[2pt]
Avoidance & $\mathbf{F}(\ell^{a}_{5}\wedge\mathbf{F}\ell^{b}_{5})\wedge\mathbf{G}\neg\ell^{c}_{0}$ \\[2pt]
Conjunctive & $\mathbf{F}(\ell^{a}_{5}\wedge\mathbf{F}\ell^{b}_{5})\wedge\mathbf{F}(\ell^{c}_{5}\wedge\mathbf{F}\ell^{d}_{5})$ \\[2pt]
Disjunctive & $\bigl(\mathbf{F}(\ell^{a}_{5}\wedge\mathbf{F}\ell^{b}_{5})\wedge\mathbf{G}\neg\ell^{c}_{0}\bigr)
              \vee \bigl(\mathbf{F}(\ell^{c}_{5}\wedge\mathbf{F}\ell^{d}_{5})\wedge\mathbf{G}\neg\ell^{a}_{0}\bigr)$ \\
\bottomrule
\end{tabular}
\vspace{-10pt}
\end{table}

\subsection{Metrics}
\label{sec:metrics}

We measure the success rate (\textbf{SR}) for the mission completion as the fraction of episodes reaching an accepting terminal state. Efficiency is measured by the Time To Satisfaction (\textbf{TTS}) from the first motion measured in seconds, and the distance (\textbf{Dist.}) traversed by the resulting plan, measured in meters. We also measure coverage (\textbf{Cov.}) as the percentage of the geometrically explored environment, indicating how much exploration effort was required.

We report failure causes separately, to analyze failure modes of the methods. An exploration failure (\textbf{Expl.}) is a run that ended without finding the mission-related objects, a sequence (\textbf{Seq.} failure happens when the required mission sequence was not respected, and an avoidance (\textbf{Avoid.}) failure is identified when a negated proposition was activated.

\subsection{Baselines}

We compare our proposed method against representative methods
for geometric exploration, semantic exploration, and LTL planning. When
necessary, we implement or adapt them in order to accept the missions specified
in our benchmark. For fairness, every baseline shares our scene graph and our
perception and navigation stacks, so that the comparison is on the planning
strategy alone.

As a reference for the performance of planning methods in unknown environments,
we add the \textbf{Known map} method, which uses the ground-truth scene graph and plans offline~\cite{kloetzer2009automatic}. This represents the gold
standard for optimal plans. We also apply two state-of-the-art exploration
strategies. \textbf{FUEL}~\cite{zhou2021fuel} performs an exhaustive geometric
exploration of the environment, whereas \textbf{VLFM}~\cite{yokoyama2024vlfm}
searches for the targets one at a time. In both cases, an LTL planning method is
applied on the resulting scene graph~\cite{kloetzer2009automatic}. We also apply
a greedy SubTask selection strategy~\cite{xu2022framework} \textbf{Greedy ST}, which
scores the exploration and satisfaction actions myopically at each planning
stage and executes the best one. Finally, we compare against a strategy that
explores guided by the object of the current mission stage until it is found,
and only then addresses the following one, named \textbf{Percep-LTL}~\cite{kantaros2022perception}.

\subsection{Experiments}
\label{sec:configs}

We also perform a sensitivity analysis on the parameters of the waypoint value
and ablate different components of our method. \emph{No hints} removes the
semantic guidance, leaving the exploration purely geometric while the VLM still
detects the targets, which measures the contribution of the mission-driven
exploration. \emph{Value composition} sweeps the value of each waypoint group
while the others are kept fixed, which measures how the balance between
exploring and advancing the mission affects the results.

\begin{table}[t]
  \centering
  \caption{Results of the benchmark including different mission types. Among the
  compared methods, the best value is shaded in \colorbox{green!22}{green} and
  the second best in \colorbox{blue!13}{blue}. \emph{Known-map}
  (\gm{grey}) is the reference gold-standard. TTS and Path are reported as mean $\pm$ standard
  deviation across episodes. SR and Cov are \%, TTS are seconds, and Dist. are meters.}
  \label{tab:main}
  \footnotesize
  \setlength{\tabcolsep}{2.5pt}
  \begin{tabular}{@{}clcccccccc@{}}
    \toprule
    & & \multicolumn{4}{c}{Outcome}
      & \multicolumn{3}{c}{Error analysis $\bm{\downarrow}$} \\
    \cmidrule(lr){3-6}\cmidrule(lr){7-9}
    & Method & SR $\bm{\uparrow}$ & TTS $\bm{\downarrow}$ & Dist. $\bm{\downarrow}$ & Cov $\bm{\downarrow}$
    & Expl. & Seq. & Avoid. \\
    \midrule

    \multirow{6}{*}{\rotatebox[origin=c]{90}{Object search}}
      & \gm{Known map~\cite{kloetzer2009automatic}} & \gm{100} & \gm{39\ppm{12}} & \gm{8.2\ppm{0.9}} & \gm{---} & \gm{---} & \gm{---} & \gm{---} \\
      & FUEL$+$P.~\cite{zhou2021fuel} & 63.9 & 338\ppm{52} & 53.2\ppm{5.2} & 93.0 & \err{13} & --- & --- \\
      & VLFM$+$P.~\cite{yokoyama2024vlfm} & \bst{83.3} & \bst{64\ppm{16}} & \snd{26.4\ppm{6.2}} & \bst{46.1} & \err{6} & --- & --- \\
      & Greedy ST~\cite{xu2022framework} & 75.0 & \snd{86\ppm{12}} & 27.3\ppm{5.6} & 56.8 & \err{9} & --- & --- \\
      & Percep-LTL~\cite{kantaros2022perception} & \bst{83.3} & 100\ppm{11} & 29.4\ppm{7.7} & 62.0 & \err{6} & --- & --- \\
      & Ours & \snd{80.6} & 93\ppm{21} & \bst{24.1\ppm{7.6}} & \snd{53.5} & \err{7} & --- & --- \\
    \midrule

    \multirow{6}{*}{\rotatebox[origin=c]{90}{Sequentiality}}
      & \gm{Known map~\cite{kloetzer2009automatic}} & \gm{100} & \gm{123\ppm{24}} & \gm{28.5\ppm{5.8}} & \gm{---} & \gm{---} & \gm{---} & \gm{---} \\
      & FUEL$+$P.~\cite{zhou2021fuel} & 55.6 & 598\ppm{41} & 72.9\ppm{8.2} & 92.8 & \err{3} & \err{13} & --- \\
      & VLFM$+$P.~\cite{yokoyama2024vlfm} & 69.4 & 301\ppm{42} & 63.5\ppm{9.2} & 75.7 & \err{3} & \err{8} & --- \\
      & Greedy ST~\cite{xu2022framework} & 72.2 & \snd{274\ppm{39}} & 59.8\ppm{6.9} & \bst{72.6} & \err{8} & \err{2} & --- \\
      & Percep-LTL~\cite{kantaros2022perception} & \bst{80.6} & 313\ppm{37} & \snd{55.6\ppm{9.2}} & 85.5 & \err{7} & \err{0} & --- \\
      & Ours & \snd{77.8} & \bst{258\ppm{42}} & \bst{51.2\ppm{11.2}} & \snd{74.9} & \err{8} & \err{0} & --- \\
    \midrule

    \multirow{6}{*}{\rotatebox[origin=c]{90}{Avoidance}}
      & \gm{Known map~\cite{kloetzer2009automatic}} & \gm{100} & \gm{58\ppm{8}} & \gm{14.9\ppm{3.9}} & \gm{---} & \gm{---} & \gm{---} & \gm{---} \\
      & FUEL$+$P.~\cite{zhou2021fuel} & 0.0 & --- & --- & 93.4 & \err{0} & \err{9} & \err{27} \\
      & VLFM$+$P.~\cite{yokoyama2024vlfm} & 19.4 & \bst{79\ppm{6}} & \bst{18.5\ppm{3.2}} & 71.6 & \err{1} & \err{9} & \err{19} \\
      & Greedy ST~\cite{xu2022framework} & 52.8 & \snd{98\ppm{26}} & 24.8\ppm{4.1} & \snd{57.8} & \err{7} & \err{2} & \err{8} \\
      & Percep-LTL~\cite{kantaros2022perception} & \snd{77.8} & 183\ppm{38} & 38.9\ppm{11.6} & 58.3 & \err{5} & \err{0} & \err{3} \\
      & Ours & \bst{80.6} & 112\ppm{24} & \snd{23.5\ppm{4.9}} & \bst{49.4} & \err{6} & \err{0} & \err{1} \\
    \midrule

    \multirow{6}{*}{\rotatebox[origin=c]{90}{Conjunction}}
      & \gm{Known map~\cite{kloetzer2009automatic}} & \gm{100} & \gm{186\ppm{45}} & \gm{36.1\ppm{7.2}} & \gm{---} & \gm{---} & \gm{---} & \gm{---} \\
      & FUEL$+$P.~\cite{zhou2021fuel} & 11.1 & 621\ppm{57} & 64.6\ppm{9.9} & 94.5 & \err{7} & \err{25} & --- \\
      & VLFM$+$P.~\cite{yokoyama2024vlfm} & 27.8 & 460\ppm{48} & 55.5\ppm{8.9} & 88.1 & \err{4} & \err{22} & --- \\
      & Greedy ST~\cite{xu2022framework} & 44.4 & \snd{293\ppm{49}} & \snd{43.4\ppm{10.4}} & \snd{82.8} & \err{16} & \err{4} & --- \\
      & Percep-LTL~\cite{kantaros2022perception} & \snd{58.3} & 332\ppm{52} & 45.8\ppm{12.8} & 84.8 & \err{15} & \err{0} & --- \\
      & Ours & \bst{63.9} & \bst{283\ppm{48}} & \bst{39.9\ppm{8.3}} & \bst{81.3} & \err{13} & \err{0} & --- \\
    \midrule

    \multirow{6}{*}{\rotatebox[origin=c]{90}{Disjunction}}
      & \gm{Known map~\cite{kloetzer2009automatic}} & \gm{100} & \gm{62\ppm{8}} & \gm{13.6\ppm{2.9}} & \gm{---} & \gm{---} & \gm{---} & \gm{---} \\
      & FUEL$+$P.~\cite{zhou2021fuel} & 0.0 & --- & --- & 93.6 & \err{0} & \err{13} & \err{23} \\
      & VLFM$+$P.~\cite{yokoyama2024vlfm} & 38.9 & 400\ppm{42} & 69.8\ppm{8.3} & 86.7 & \err{1} & \err{7} & \err{14} \\
      & Greedy ST~\cite{xu2022framework} & 55.6 & \snd{279\ppm{40}} & \snd{41.3\ppm{8.1}} & \snd{70.3} & \err{3} & \err{3} & \err{10} \\
      & Percep-LTL~\cite{kantaros2022perception} & \snd{61.1} & 293\ppm{35} & 44.7\ppm{9.2} & 76.5 & \err{11} & \err{0} & \err{3} \\
      & Ours & \bst{72.2} & \bst{245\ppm{36}} & \bst{33.5\ppm{6.4}} & \bst{63.4} & \err{9} & \err{0} & \err{1} \\
    \bottomrule
  \end{tabular}
\vspace{-10pt}
\end{table}

\section{Results}
\label{sec:results}

\subsubsection{Baseline comparison} We report the main results of the benchmark comparison in Table~\ref{tab:main} for the five mission types. For simpler missions involving the search of a single object, most methods perform comparably. VLFM stands out, since the method is specifically designed for object search. In contrast, FUEL has the lowest performance despite exploring the whole environment, indicating the advantage of a semantically-driven exploration. In sequential missions, it can already be observed that detaching the exploration from the mission execution results in violating the ordering imposed by the mission. In these scenarios, the methods considering the LTL mission at all times perform comparably.

\begin{figure*}[h!]
    \centering
    \includegraphics[width=0.85\linewidth]{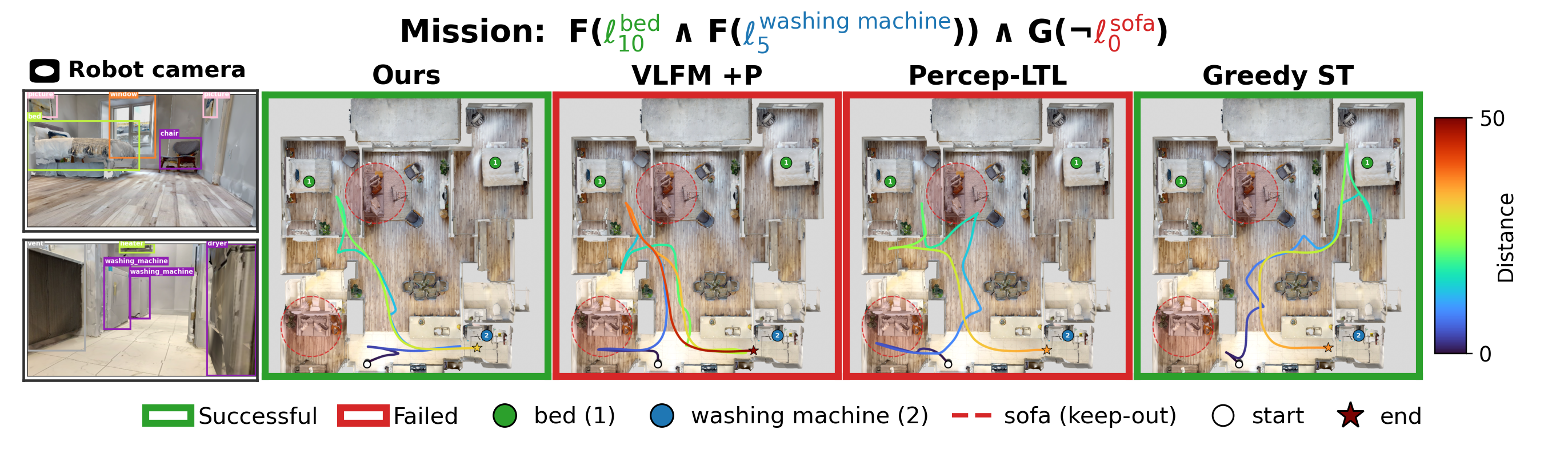}
    \caption{VLFM$+$P.~\cite{yokoyama2024vlfm} explores until all the objects are found,
  which lengthens the path and violates a restriction while exploring. Percep-LTL~\cite{kantaros2022perception} also explores first, and the lack of
  semantic guidance induces an avoidance error. Greedy
  ST~\cite{xu2022framework} reaches the first object without any information
  about the next one. Ours reaches both in order and keeps clear of the sofa,
  as the semantic guidance provides early evidence of the second target and
  yields the shortest path.}
    \label{fig:results}
    \vspace{-14pt}
\end{figure*}

In more complex missions involving avoidance and branching, the performance of the baselines that split the mission into two stages drops significantly, mainly due to violating the mission restrictions. Fig.~\ref{fig:results} shows a visual example of the solutions generated by the different baselines for one mission, where it can be seen that our method obtains the shortest path. The methods considering the LTL mission at all times retain their performance, although some differences appear. Percep-LTL searches for the object it currently needs, which reduces its efficiency. On the other hand, every method except ours and Percep-LTL, the two non-myopic ones, incurs in sequence failures. The
further error analysis shows that most of our failure cases are on object search, mainly related to perception. The few avoidance failures are attributed to not being able to avoid the objects that are not yet registered in the scene graph.

Overall, our method achieves the highest success rate for almost every mission type, specially in the complex ones, with almost no errors related to the mission execution. Meanwhile, it obtains the most efficient paths in these complex missions, while exploring the least amount of the environment. The improvement in efficiency indicates the advantage of interleaving exploration with mission execution in complex missions, while the reduced observed environment showcases the advantages of considering both the semantic guidance and the mission information for information acquisition.

\begin{table}[t]
  \centering
  \small
  \caption{Ablation of the semantic hints. \emph{No hints} disables the language
    model and builds no semantic waypoints, leaving pure frontier exploration,
    the VLM still detects targets. Fifteen episodes per mission type over five
    scenes.}
    \label{tab:norel_ablation}
  \setlength{\tabcolsep}{4pt}
  \begin{tabular}{@{}llcccc@{}}
    \toprule
    & Mission & SR (\%) & TTS (s) & Dist. (m) & Cov (\%) \\
    \midrule
    \multirow{3}{*}{\rotatebox[origin=c]{90}{\emph{Ours}}}
      & Obj.\ search   & 100.0 & 102.1$\pm$12.4 & 22.5$\pm$4.8  & 44.3 \\
      & Conj.\ + avoid &  80.0 & 200.9$\pm$18.6 & 46.5$\pm$7.2  & 67.7 \\
      & Sequential     &  60.0 & 291.4$\pm$22.7 & 65.8$\pm$9.4  & 82.0 \\
    \midrule
    \multirow{3}{*}{\rotatebox[origin=c]{90}{\emph{No hints}}}
      & Obj.\ search   & 100.0 & 177.4$\pm$21.2 & 33.5$\pm$7.6  & 69.8 \\
      & Conj.\ + avoid &  53.3 & 357.1$\pm$34.5 & 62.4$\pm$10.1 & 94.2 \\
      & Sequential     &  33.3 & 439.8$\pm$40.3 & 82.8$\pm$11.7 & 98.5 \\
    \bottomrule
  \end{tabular}
\vspace{-10pt}
\end{table}

\subsubsection{Ablation}
Table~\ref{tab:norel_ablation} isolates the contribution of the semantic hints. Removing them leaves the same mission reasoning with a purely geometric exploration, and the robot covers far more of the scene to complete fewer missions, taking longer and traveling further. The effect grows with the number of objects a mission requires. In object search, both variants always succeed, but the guided search finds the target sooner, whereas once several objects have to be located, the success rate drops without hints. These results show that a mission-driven semantic search makes the information acquisition more effective and efficient.

\begin{table}[t]
  \centering
  \small
  \caption{Value sensitivity. Each group is swept while the others stay frozen. The tuned
    operating point is given for reference. Twelve episodes of the nested
    sequential mission, over four scenes.}
    \label{tab:value_ablation_results}
  \setlength{\tabcolsep}{4pt}
  \begin{tabular}{@{}lccccc@{}}
    \toprule
    Group & $v$ & SR (\%) & TTS (s) & Dist. (m) & Cov (\%) \\
    \midrule
    Tuned & ---  & \textbf{75.0} & \textbf{234.9$\pm$22.1} & \textbf{47.6$\pm$6.3} & \textbf{68.3} \\
    \midrule
    \multirow{4}{*}{Mission}
          & 0.00 & 50.0 & 389.4$\pm$35.8 & 59.6$\pm$7.4 & 95.3 \\
          & 0.25 & 58.3 & 359.9$\pm$31.2 & 57.2$\pm$8.2 & 91.2 \\
          & 0.50 & 66.7 & 288.7$\pm$27.5 & 52.3$\pm$6.8  & 77.6 \\
          & 0.75 & 75.0 & 255.6$\pm$22.4 & 49.1$\pm$6.2  & 71.5 \\
    \midrule
    \multirow{4}{*}{Semantic}
          & 0.00 & 75.0 & 422.1$\pm$42.7 & 61.9$\pm$8.1 & 96.2 \\
          & 0.25 & 66.7 & 362.3$\pm$34.6 & 57.8$\pm$6.9 & 93.2 \\
          & 0.50 & 75.0 & 284.9$\pm$27.8 & 60.2$\pm$7.2 & 86.4 \\
          & 0.75 & 83.3 & 267.2$\pm$23.7 & 52.9$\pm$6.6  & 78.2 \\
    \midrule
    \multirow{4}{*}{Geometric}
          & 0.25 & 83.3 & 282.6$\pm$24.9 & 53.5$\pm$9.7  & 77.5 \\
          & 0.50 & 75.0 & 299.3$\pm$23.4 & 56.1$\pm$8.5 & 82.4 \\
          & 0.75 & 66.7 & 317.8$\pm$28.1 & 60.9$\pm$7.0 & 84.3 \\
          & 1.00 & 75.0 & 375.2$\pm$36.3 & 66.8$\pm$7.8 & 92.1 \\
    \bottomrule
  \end{tabular}

\end{table}

\subsubsection{Sensitivity analysis}

Table~\ref{tab:value_ablation_results} sweeps the value of each waypoint group while the others are kept fixed. Lowering the mission value makes the robot explore when it could already advance the mission, which increases the coverage and reduces the success rate. When lowering the semantic value, the robot explores geometrically, maintaining SR but impacting the efficiency as the missions are still completed but require much longer trajectories. Raising the geometric value has a similar effect, since the frontiers start competing with the semantic evidence. Confirming the correctness of the value ordering proposed in our method: reaching a proposition is worth more than following a semantic hint, and a hint more than a frontier.

\section{Conclusions}
\label{sec:conclusion}

This paper presented a novel adaptive high-level planning method to solve LTLf missions in unknown environments. Our method interleaves a task-driven semantic search with the execution of the mission, leveraging a metric-semantic scene graph built online with a VLM and the DFA encoding the mission. At every planning stage, our planner selects the waypoints that are most valuable for both the semantic search and the advancement of the mission, reasoning over the remaining stages in order to avoid blocking states. The selected waypoints are then ordered in a single high-level plan, which is recomputed as new information arrives. Extensive experiments in a photorealistic simulator show that our method achieves a higher success rate in the most complex missions, while being more efficient in mission time and traversed
distance. Our ablations show that each component contributes differently, and that the best performance is obtained when they are combined. Future work will study the extension of our approach to several robots sharing one mission, where the information about the environment has to be shared and the planner has to
consider the requirements and restrictions of the whole team.
 
\balance

\end{document}